\documentclass{article} 
\usepackage{iclr2026_conference,times}

\usepackage{amsmath,amsfonts,bm}

\def\eqref#1{equation~\ref{#1}}

\def\1{\bm{1}}

\DeclareMathAlphabet{\mathsfit}{\encodingdefault}{\sfdefault}{m}{sl}
\SetMathAlphabet{\mathsfit}{bold}{\encodingdefault}{\sfdefault}{bx}{n}

\usepackage{hyperref}
\usepackage{url}
\usepackage{booktabs}
\usepackage{graphicx}
\usepackage{multirow}
\usepackage[table]{xcolor}
\definecolor{rowhl}{RGB}{232,242,255}
\usepackage{tcolorbox}
\tcbuselibrary{listings,skins}
\makeatletter
\renewcommand\paragraph{\@startsection{paragraph}{4}{\z@}{0.5ex plus 0.2ex minus 0.1ex}{-1em}{\normalfont\normalsize\bfseries}}
\makeatother
\newtcolorbox{promptbox}[1][]{
  enhanced, colback=gray!5, colframe=gray!60,
  fonttitle=\bfseries\small, title=#1,
  boxrule=0.5pt, arc=2pt, left=6pt, right=6pt, top=4pt, bottom=4pt,
  fontupper=\small\ttfamily
}

\title{Track, Rank, Crack: Epistemic Working Memory Scales Multi-Hop Reasoning in Language Agents}

\author{Ning Liu \\
Independent Researcher \\
\texttt{ningliu@umich.edu}
}

\iclrpreprintcopy
\begin{document}

\maketitle

\begin{abstract}

Language agents that interleave reasoning and tool use degrade sharply as reasoning chains lengthen, even when each individual step is easy. We trace this to \emph{context dilution}: an agent's investigative state (what it has confirmed, what it suspects, and what it still needs) lives only implicitly in a growing context window, where early discoveries are buried under later retrievals. We introduce SLEUTH, which makes this state explicit and actionable through a structured epistemic working memory: the agent maintains Confirmed Facts grounded to sources, Active Hypotheses ranked by evidence, and Open Questions that directly drive its next action. Across five multi-hop benchmarks and five established baselines, SLEUTH's advantage grows with difficulty, from +5 points on HotpotQA to +11 on 4-hop chains, surpassing Reflexion without multiple episodes. Analyzing where the remaining gap lies, we identify the \emph{evidence sufficiency problem}: agents often find the answer but fail to commit, exhausting their budget on needless verification. A lightweight commitment trigger fixes this, but only when the agent already maintains structured state: the identical trigger applied to an unstructured agent yields no improvement, isolating organized epistemic state as the necessary condition for effective commitment. Finally, enforcing protocol adherence on a weaker model recovers up to +19 points on the hardest problems, showing that \emph{how} an agent organizes its reasoning, not raw model capability, is the active ingredient for scaling multi-hop reasoning.

\end{abstract}

\section{Introduction}
\label{sec:intro}

Effective reasoning requires not only the ability to draw inferences but also the ability to manage what is known, what is uncertain, and what remains to be determined~\citep{baddeley1992working,baddeley2000episodic,newell1972human}. Language model agents that interleave reasoning with tool use~\citep{yao2023react,schick2024toolformer} have made striking progress on multi-hop question answering~\citep{yang2018hotpotqa,trivedi2022musique,ho2020constructing}, yet they exhibit a characteristic failure as reasoning chains grow: agents that reliably answer 2-hop questions degrade sharply on 3- and 4-hop problems, even when the constituent sub-questions are individually straightforward. Why does composing more reasoning steps cause such disproportionate difficulty?

We argue that the root cause is \emph{context dilution}---the progressive burial of decision-relevant information under accumulating tool outputs. While retrieval-augmented generation~\citep{lewis2020retrieval,asai2024selfrag} has made single-hop retrieval effective, multi-hop problems compound the challenge: a ReAct agent~\citep{yao2023react} must attend over dozens of retrieved paragraphs while simultaneously tracking which sub-questions are resolved, which hypotheses remain viable, and what to investigate next. This investigative state is never made explicit; it exists only implicitly in the growing context window, where early discoveries compete for attention with later retrievals. The longer the chain, the more severe the dilution.

Human investigators---detectives, diagnosticians, scientists---solve precisely this problem by maintaining structured working notes that separate \emph{what is known} from \emph{what is hypothesized} from \emph{what remains to be determined}. This tripartite organization serves a dual purpose: it compresses the investigation state into a decision-relevant summary, and it directly determines the next action (investigate the highest-priority open question). We operationalize this insight for language agents.

We introduce \textbf{SLEUTH} (\textbf{S}tructured \textbf{L}ayered \textbf{E}vidence-\textbf{U}pdated \textbf{T}racking of \textbf{H}ypotheses), a working memory framework, implemented entirely through prompting, in which the agent maintains three typed components after every action: \textbf{Confirmed Facts} grounded to specific sources, \textbf{Active Hypotheses} ranked by evidentiary support, and \textbf{Open Questions} that drive action selection. By outputting and updating this structure on every turn, the agent carries an explicit epistemic state across the trajectory rather than leaving it implicit in the context window. This sets SLEUTH apart from memory systems that accumulate experience \emph{across} episodes~\citep{shinn2023reflexion,majumder2023clin,zhao2023expel} and from methods that scale test-time compute by searching \emph{more} trajectories~\citep{yao2023tree,zhou2024lats,snell2025scaling}: SLEUTH instead organizes information \emph{within} a single trajectory, requiring no extra episodes, search, or training. Our key contributions are:
\begin{itemize}
    \item We introduce SLEUTH, which makes an agent's epistemic state \emph{explicit and actionable}: confirmed facts, ranked hypotheses, and open questions that directly drive the next action. Unlike approaches that externalize state through code execution, symbolic controllers, or training, SLEUTH requires only prompting.
    \item We show that this structured state yields gains that \emph{scale with reasoning difficulty}---widening from +4.7 points on HotpotQA to +10.9 on 4-hop chains over five established baselines across five benchmarks---and surpasses Reflexion without multiple episodes.
    \item We identify the \emph{evidence sufficiency problem}: agents gather enough evidence but fail to commit under bounded computation. Our commitment mechanism resolves it, but only when coupled with structured state---the identical trigger that lifts SLEUTH by double digits leaves an unstructured agent unchanged, isolating organized epistemic state as the necessary condition.
    \item We establish that \emph{protocol adherence}, not model capability, is the active ingredient: enforcing consistent working-memory maintenance on a weaker model (GLM-5) recovers up to +19 points on the hardest problems.
\end{itemize}

\section{Related Work}
\label{sec:related}

\paragraph{Agent Reasoning and Multi-Hop Retrieval.}

ReAct~\citep{yao2023react} established the interleaved thought-action paradigm, but leaves the agent's epistemic state implicit in the growing context. Chain-of-Thought~\citep{wei2022chain} and its variants~\citep{wang2023selfconsistency,khot2023decomposed,zhou2023leasttomost} improve single-turn reasoning but do not track beliefs across turns. Self-Ask~\citep{press2023selfask} decomposes questions sequentially without persistent state; IRCoT~\citep{trivedi2023ircot} interleaves retrieval with chain-of-thought but represents neither what is established nor what remains open. Recent work trains iterative retrieval-reasoning policies~\citep{wang2025corag,li2025searcho1,jin2025searchr1} or adds multi-agent backtracking~\citep{zhang2025reagent}, improving retrieval integration without an explicit epistemic state. Reflexion~\citep{shinn2023reflexion} adds post-episode self-reflection but requires multiple episodes; Tree of Thoughts~\citep{yao2023tree} and LATS~\citep{zhou2024lats} scale test-time compute through trajectory search. SLEUTH is complementary, improving reasoning quality within a single trajectory.

\paragraph{Memory and State Management for Agents.}

CoALA~\citep{sumers2024coala} provides cognitive-architecture scaffolding with modular memory types; MemGPT~\citep{packer2023memgpt} manages context through OS-inspired paging. CLIN~\citep{majumder2023clin}, ExpeL~\citep{zhao2023expel}, and A-Mem~\citep{xu2025amem} learn or evolve \emph{cross-episode} knowledge, whereas SLEUTH maintains \emph{prospective} within-episode state. HiAgent~\citep{hu2025hiagent} compresses subgoal-level information; \citet{wang2024symbolic} add neurosymbolic working memory for deductive reasoning; CAT~\citep{zhang2025cat} exposes context maintenance as a callable tool. Programmatic and graph-structured controllers~\citep{besta2025demystifying} externalize state with stronger guarantees but require code execution, training, or auxiliary infrastructure. SLEUTH achieves substantial gains at the lightest design point---prompting alone---distinguished by its tripartite \emph{epistemic} structure and the direct coupling between open questions and action selection.

\paragraph{Test-Time Compute and Stopping Criteria.}

Adaptively allocating inference-time computation can match or exceed scaling model parameters~\citep{snell2025scaling,muennighoff2025s1,deepseek2025r1}, but these approaches operate on single-turn reasoning; the analogous multi-turn challenge is allocating a fixed \emph{action budget} across evidence gathering and commitment. RL-based methods~\citep{jin2025searchr1,shi2025searchrefine} train models to decide \emph{when} to search but not how to organize gathered evidence. Our nudge addresses the complementary exploration-exploitation tradeoff at the trajectory level, triggering commitment after a principled fraction (70\%) of the turn budget. Crucially, SLEUTH requires no reinforcement learning or fine-tuning---it is implemented entirely through prompting and a lightweight runtime check, making it immediately applicable to any instruction-following model.

\section{SLEUTH: Structured Epistemic Working Memory}
\label{sec:method}

\begin{figure}[t]
\centering
\includegraphics[width=\textwidth]{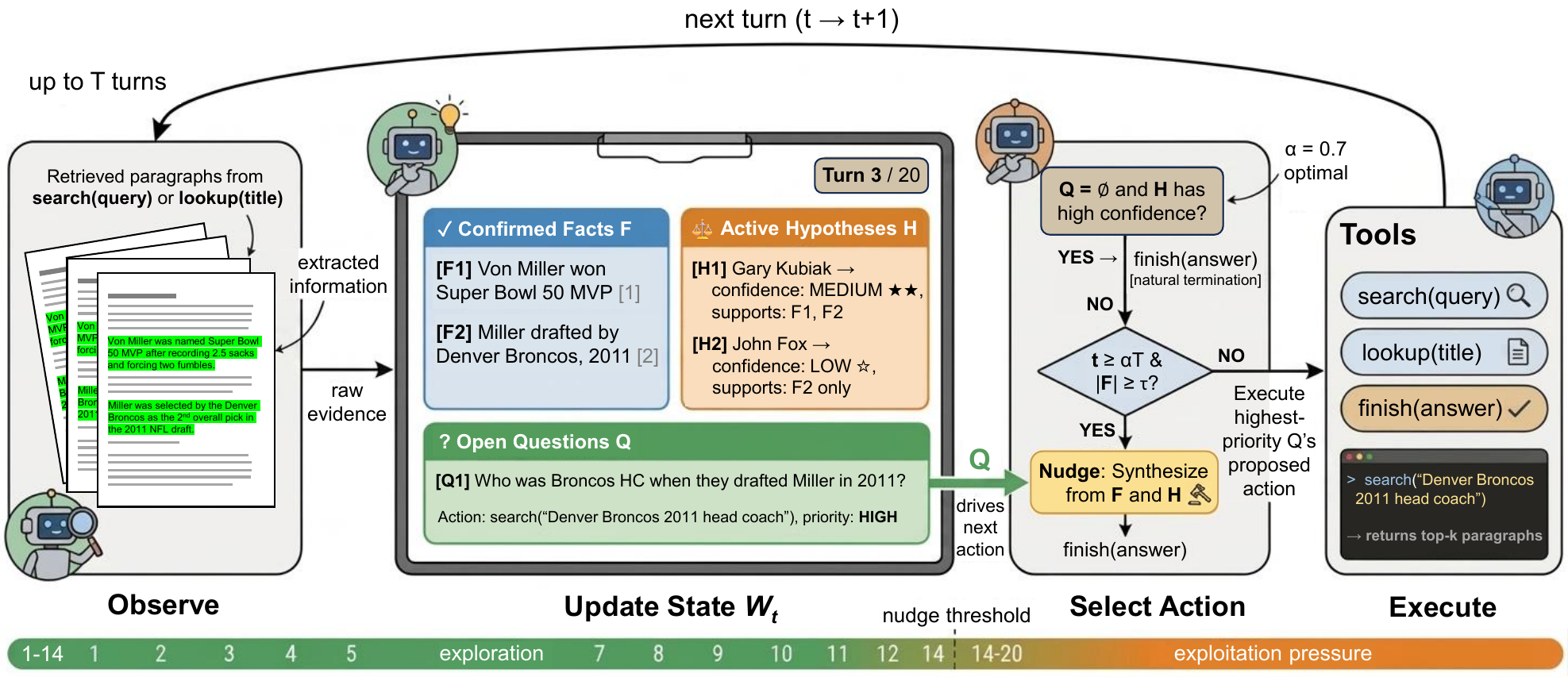}
\caption{Overview of the SLEUTH framework on a 3-hop question (shown at turn 3 of 20). The agent observes retrieved paragraphs, updates its tripartite working memory $W_t = (\mathcal{F}_t, \mathcal{H}_t, \mathcal{Q}_t)$, and selects its next action based on the highest-priority open question. The commitment mechanism (nudge) triggers at turn $\alpha T$ to force exploitation of accumulated evidence. The timeline below shows the exploration--exploitation tradeoff across the turn budget.}
\label{fig:method}
\end{figure}

SLEUTH augments a language agent with an explicit \emph{epistemic working memory}---a structured scratchpad~\citep{nye2021scratchpad} that the agent reads and updates on every turn, organized as a tripartite epistemic structure that separates established knowledge from hypotheses and open uncertainties. The framework requires no fine-tuning or architectural modification; it is implemented entirely through prompting. We describe the working memory structure (\S\ref{sec:wm-structure}), the update protocol (\S\ref{sec:update-protocol}), and the commitment mechanism that addresses the evidence sufficiency problem (\S\ref{sec:commitment}). Figure~\ref{fig:method} illustrates the framework on a multi-hop question.

\subsection{Working Memory Structure}
\label{sec:wm-structure}

At each turn $t$, the agent maintains a working memory state $W_t$ comprising three typed components:

\paragraph{Confirmed Facts} $\mathcal{F}_t = \{f_1, \ldots, f_n\}$ are verified statements extracted from tool outputs, each grounded to a specific source. Entry format: \texttt{[F$i$] $\langle$statement$\rangle$ (source: $\langle$title$\rangle$)}. Facts are monotonically accumulated---once confirmed, they are never retracted within an episode. This monotonicity ensures that evidence is preserved rather than lost to context dilution.

\paragraph{Active Hypotheses} $\mathcal{H}_t = \{h_1, \ldots, h_m\}$ are candidate answers ranked by evidentiary support. Each hypothesis is linked to the confirmed facts that support or contradict it, with an explicit confidence level (high/medium/low). Entry format: \texttt{[H$j$] $\langle$hypothesis$\rangle$ | confidence: $\langle$level$\rangle$ | supports: $\langle$fact IDs$\rangle$ | contradicts: $\langle$fact IDs$\rangle$}. Rankings change as new evidence arrives: hypotheses gain confidence when supporting facts accumulate and lose confidence when contradicted.

\paragraph{Open Questions} $\mathcal{Q}_t = \{q_1, \ldots, q_l\}$ are uncertainties whose resolution would most change the hypothesis rankings, each with a proposed action. Entry format: \texttt{[Q$k$] $\langle$question$\rangle$ | action: $\langle$proposed query$\rangle$ | priority: $\langle$level$\rangle$}. The highest-priority open question directly determines the agent's next tool call, creating a tight coupling between epistemic state and action selection. When $\mathcal{Q}_t = \emptyset$ and some $h_j$ has high confidence, the agent commits to an answer.

The complete state $W_t = (\mathcal{F}_t, \mathcal{H}_t, \mathcal{Q}_t)$ is output in full on every turn, creating an explicit chain of epistemic states that constitutes the agent's reasoning trace.

\subsection{Update Protocol}
\label{sec:update-protocol}

After every tool call, the agent performs a structured update $W_{t-1} \rightarrow W_t$:

\begin{enumerate}
    \item \textbf{Extract facts.} Parse the tool output for new verifiable information. Add each as a confirmed fact with a source pointer: $\mathcal{F}_t \supseteq \mathcal{F}_{t-1}$.
    \item \textbf{Update hypotheses.} For each $h_j \in \mathcal{H}_{t-1}$, determine whether new facts support, contradict, or are irrelevant to it. Adjust confidence accordingly. Generate new hypotheses if evidence suggests explanations not yet considered.
    \item \textbf{Update questions.} Remove questions answered by the new observation. Generate new questions targeting remaining uncertainty---specifically, questions whose answers would most change hypothesis rankings.
    \item \textbf{Select action.} Execute the highest-priority question's proposed action. If $\mathcal{Q}_t = \emptyset$ and some hypothesis has high confidence, call \texttt{finish(answer)}.
\end{enumerate}

This protocol enforces two key properties. First, \emph{evidence monotonicity}: confirmed facts never disappear, preventing the context dilution that degrades standard agents. Second, \emph{action grounding}: every tool call is justified by an explicit open question, eliminating redundant or unfocused searches. Empirically, facts accumulate from ${\sim}2$ to ${\sim}4$ while open questions decrease from ${\sim}1.5$ to ${\sim}1.1$ across a typical 4-hop trajectory (Appendix~\ref{app:wm-dynamics}). The working memory is enforced at runtime: the agent's response must contain all three sections in prescribed notation before a tool call is executed.

\subsection{Commitment Mechanism}
\label{sec:commitment}

A structured working memory improves evidence organization but introduces a new failure mode: \emph{over-verification}. The hypothesis-confidence framework, while useful for ranking alternatives, provides no clear stopping criterion. We observe empirically that agents maintaining high standards for ``sufficient evidence'' continue gathering confirmatory evidence even after the answer is determinable, exhausting their turn budget on questions that have already been effectively resolved.

We address this with a \emph{nudge} mechanism---a commitment trigger that fires when two conditions are jointly satisfied: (1)~the agent has consumed at least fraction $\alpha$ of its turn budget $T$ (i.e., turn $t \geq \alpha T$), and (2)~the agent has accumulated at least $\tau$ confirmed facts ($|\mathcal{F}_t| \geq \tau$), ensuring there is enough evidence to synthesize from. We set $\tau{=}2$, which we find works well empirically: it is the minimum needed to compose a multi-hop answer while still gating out premature commitment when the agent has barely begun gathering evidence. The trigger is deliberately independent of the agent's open-question state: in multi-hop problems, resolving one sub-question surfaces the next link in the chain, so $\mathcal{Q}_t$ is legitimately non-empty until the reasoning chain completes---conditioning on $\mathcal{Q}_t = \emptyset$ costs ${\sim}$4 EM points as the nudge effectively never fires. When triggered, the agent receives a synthesis message instructing it to construct an answer from its confirmed facts and active hypotheses without further search. This transforms the open-ended investigation into a commitment decision, forcing the agent to exploit its accumulated evidence rather than continuing to explore. The threshold $\alpha$ controls an exploration-exploitation tradeoff: too small (e.g., $\alpha{=}0.3$) cuts off investigation before sufficient evidence is gathered; too large (e.g., $\alpha{=}0.8$) leaves insufficient margin for synthesis. We find $\alpha{=}0.7$ optimal empirically (\S\ref{sec:nudge-ablation}).

\subsection{Implementation}
\label{sec:implementation}

SLEUTH is implemented as a system prompt that instructs the agent to maintain the working memory structure described above. The agent is given three tools: \texttt{search(query)}, which returns the top-$k$ most relevant paragraphs from the question's context; \texttt{lookup(title)}, which returns the full text of a specific paragraph; and \texttt{finish(answer)}, which terminates the episode with a final answer. The same tool interface is used across all baselines; only the system prompt differs.

For models with weaker instruction-following, working memory adherence can be enforced through two mechanisms: (1)~prompt adaptation with explicit structural delimiters that make the format unambiguous, and (2)~a runtime check that rejects responses missing proper notation and re-prompts the agent. Together, these raise adherence from ${\sim}$15\% to ${\sim}$99\% of turns on GLM-5, recovering over 19 points on the hardest problems (\S\ref{sec:adherence}). Sonnet requires neither mechanism---it maintains the protocol through instruction-following alone.

\section{Experimental Setup}
\label{sec:experiments}

We evaluate SLEUTH against five established baselines, together with a set of working-memory ablations, on five multi-hop reasoning benchmarks spanning 2- to 4-hop reasoning chains, using two model families.

\paragraph{Datasets.} We evaluate on three multi-hop QA benchmarks: HotpotQA~\citep{yang2018hotpotqa} ($N{=}7{,}405$, 2-hop, distractor setting with 10 paragraphs per question), MuSiQue~\citep{trivedi2022musique} (2/3/4-hop answerable splits with $N{=}1{,}252/760/405$ and 20 paragraphs, enabling controlled difficulty analysis), and 2WikiMultiHopQA~\citep{ho2020constructing} ($N{=}12{,}576$, comparison and bridge reasoning). All use the validation split with full paragraph context provided to the retrieval toolkit. Full dataset statistics are in Appendix~\ref{app:datasets}.

\paragraph{Baselines.} We compare against five established agent reasoning paradigms: \textbf{ReAct}~\citep{yao2023react} interleaves reasoning and acting with no persistent state; \textbf{ReAct-CoT} augments ReAct with explicit chain-of-thought before each action; \textbf{Self-Ask}~\citep{press2023selfask} decomposes questions into sub-questions answered sequentially; \textbf{IRCoT}~\citep{trivedi2023ircot} interleaves retrieval with chain-of-thought; and \textbf{Reflexion}~\citep{shinn2023reflexion} adds post-episode self-reflection over 2 episodes.

\paragraph{Working-memory ablations.} To isolate the contribution of each working-memory component, we additionally evaluate variants of SLEUTH itself. \textbf{Notes} augments ReAct with an unstructured scratchpad, testing whether \emph{any} persistent state helps. \textbf{Facts-Only} maintains structured fact extraction without hypotheses or questions; \textbf{Facts+Hyp.} adds ranked hypotheses; and \textbf{Facts+Q.} instead adds open questions (omitting hypotheses). SLEUTH completes the tripartite structure with both. These two-component variants test whether either added component alone suffices (schema sensitivity, \S\ref{app:schema}). All methods share the same tool interface; only the system prompt differs.

\paragraph{Models and configuration.} Primary experiments use Claude Sonnet 4.6; cross-model experiments use GLM-5. All agents operate with a 20-turn budget and search top-$k{=}3$ paragraphs. We report Exact Match (EM) after standard normalization (lowercasing, article removal, whitespace). For analysis, we decompose EM into \emph{conditional accuracy} (EM among questions answered within the turn budget) and \emph{timeout rate} (fraction exhausting all turns without calling \texttt{finish}). F1 scores are reported in Appendix~\ref{app:f1}.

\section{Results}
\label{sec:results}

\subsection{Main Results}

Table~\ref{tab:main} presents exact match scores across all five benchmarks. SLEUTH achieves the highest score on every dataset, with improvements over external baselines that scale with difficulty. On HotpotQA (2-hop), SLEUTH achieves 66.1\% EM, surpassing Reflexion (61.4\%) by 4.7 points \emph{without requiring multiple episodes}. The advantage grows on MuSiQue: +11.9 points on 2-hop, +10.4 on 3-hop, and +10.9 on 4-hop over the strongest external baseline. On 2WikiMultiHopQA, SLEUTH achieves 76.4\%, a 2.8-point improvement over Reflexion.

\paragraph{What drives the gain.} The middle block isolates each working-memory component, all sharing SLEUTH's commitment and verbatim-extraction rules. Maintaining sourced facts alone (Facts-Only) already recovers most of the advantage---49.9\% on 4-hop, +7.9 over ReAct---confirming that preventing context dilution is the dominant effect. Adding hypotheses \emph{or} open questions in isolation does not help further (Facts+Hyp.\ 48.9\%, Facts+Q.\ 47.7\% on 4-hop); only the full tripartite structure adds a further +3.2 points, because open questions become actionable only once ranked hypotheses give them candidates to discriminate (\S\ref{app:schema}). Notes---an unstructured scratchpad---provides inconsistent benefit, confirming that \emph{structured} state, not merely persistent state, is the active ingredient.

\begin{table}[t]
\caption{Exact match (\%) across five multi-hop reasoning benchmarks. Best result in \textbf{bold}. The top block lists established baselines; the middle block lists working-memory ablations of SLEUTH (\S\ref{app:schema}). All methods use Claude Sonnet 4.6 with 20-turn budget and top-$k{=}3$ retrieval. $\Delta$ shows improvement over the strongest \emph{external} baseline per dataset.}
\label{tab:main}
\centering
\small
\begin{tabular}{lccccc}
\toprule
\textbf{Method} & \textbf{HotpotQA} & \textbf{2-hop} & \textbf{3-hop} & \textbf{4-hop} & \textbf{2Wiki} \\
\midrule
ReAct & 60.2 & 54.6 & 52.0 & 42.0 & 73.1 \\
ReAct-CoT & 57.7 & 53.4 & 51.7 & 42.2 & 71.8 \\
Self-Ask & 58.8 & 51.8 & 48.7 & 38.5 & 72.0 \\
IRCoT & 56.8 & 52.6 & 50.4 & 39.0 & 70.8 \\
Reflexion & 61.4 & 54.2 & 53.3 & 42.2 & 73.6 \\
\midrule
Notes & 57.6 & 54.5 & 53.0 & 41.7 & 71.8 \\
Facts-Only & 62.3 & 61.6 & 60.5 & 49.9 & 74.5 \\
Facts+Q. & 60.9 & 61.3 & 59.1 & 47.7 & 74.0 \\
Facts+Hyp. & 62.6 & 61.3 & 60.1 & 48.9 & 75.0 \\
\midrule
\rowcolor{rowhl}
\textbf{SLEUTH} & \textbf{66.1} & \textbf{66.5} & \textbf{63.7} & \textbf{53.1} & \textbf{76.4} \\
\midrule
\rowcolor{rowhl}
$\Delta$ vs.\ best ext.\ baseline & +4.7 & +11.9 & +10.4 & +10.9 & +2.8 \\
\bottomrule
\end{tabular}
\end{table}

\subsection{Scaling with Reasoning Difficulty}
\label{sec:scaling}

\begin{figure}[t]
\centering
\includegraphics[width=\linewidth]{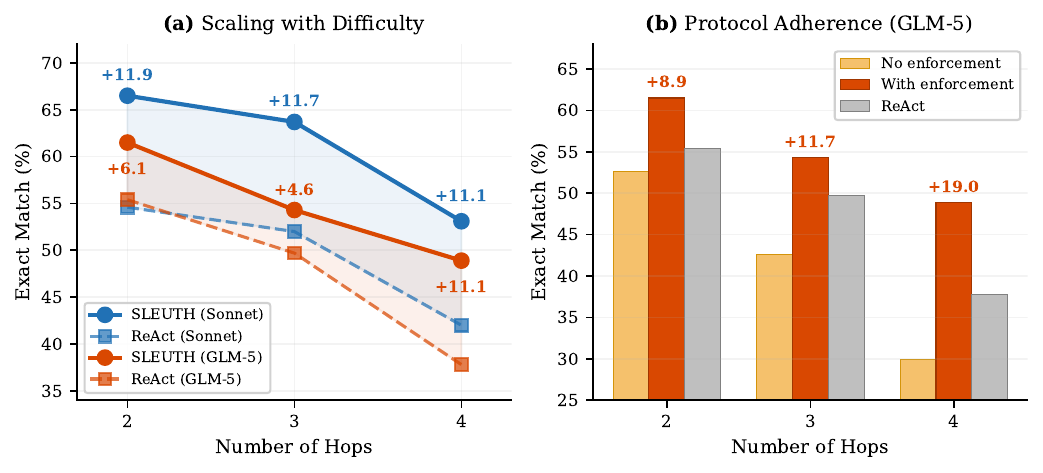}
\caption{(a) Exact match vs.\ reasoning depth on MuSiQue. SLEUTH's advantage over ReAct is consistent across both Sonnet (+11.1 to +11.9) and GLM-5 (+4.6 to +11.1), with gains largest on 4-hop where unstructured agents suffer high timeout rates. Deltas computed from unrounded scores. (b) Effect of protocol adherence on GLM-5 across MuSiQue splits: ensuring adherence through prompt adaptation and runtime validation recovers +8.9 to +19.0 points, scaling with difficulty.}
\label{fig:scaling}
\end{figure}

The MuSiQue splits enable controlled comparison across reasoning depth. Figure~\ref{fig:scaling}(a) plots EM against hop count for both model families. All agents degrade as chains lengthen, but SLEUTH degrades more gracefully: its margin over ReAct holds near +11 points across 2-, 3-, and 4-hop on Sonnet, and widens to +11.1 on 4-hop with GLM-5. The advantage is largest precisely where implicit state tracking breaks down---on the longest chains, where unstructured agents lose track of intermediate findings and exhaust their budget (timeout dynamics analyzed in \S\ref{sec:evidence-sufficiency}). Structured working memory becomes more valuable as the reasoning it must support grows harder. All SLEUTH--ReAct gaps are significant at $p<0.001$ by a paired bootstrap, and the scaling trend is robust: even the lower bound of the 4-hop confidence interval exceeds the HotpotQA point estimate (Appendix~\ref{app:significance}).

\subsection{The Evidence Sufficiency Problem and Commitment Mechanism}
\label{sec:evidence-sufficiency}
\label{sec:nudge-ablation}

Why does SLEUTH still fall short on 4-hop questions? We decompose EM into \emph{conditional accuracy} (accuracy given an answer is produced) and \emph{timeout rate} (fraction exhausting the 20-turn budget without committing). The decomposition is revealing: SLEUTH without the commitment mechanism reasons far better than ReAct (conditional accuracy 64.6\% vs.\ 51.1\%) yet squanders this on a 24.7\% timeout rate, and 92\% of those timed-out episodes contain the gold answer in their history. The agent \emph{finds} the answer but fails to \emph{commit}. We term this the \emph{evidence sufficiency problem}: the hypothesis-confidence framework that improves reasoning provides no principled stopping criterion, so the agent's evidence standard recedes continually and it over-verifies even when the answer is effectively determined.

We characterize the resulting exploration--exploitation tradeoff by sweeping the commitment trigger across the turn budget (Table~\ref{tab:nudge}). EM peaks and plateaus in the 50--70\% range, indicating robustness to exact timing. The two extremes show the tension directly: early nudges (Nudge@6) force premature commitment and depress conditional accuracy, while no nudge attains the highest conditional accuracy (64.6\%) but bleeds it away to timeouts. Nudge@14 ($\alpha{=}0.7$) is the sweet spot---it converts SLEUTH's superior reasoning into superior task performance, lifting EM from 49.1\% to 53.1\%. Accuracy-at-turn curves (Appendix~\ref{app:convergence}) confirm the mechanism: SLEUTH and ReAct track closely through turns 12--14, diverging sharply only once the nudge fires.

\begin{table}[t]
\caption{Effect of commitment timing on 4-hop MuSiQue. \emph{Nudge@$N$} triggers commitment at turn $N$ of 20. Conditional accuracy (Cond.\ Acc.) measures reasoning quality independent of timeout; timeout rate measures exploration cost.}
\label{tab:nudge}
\centering
\small
\begin{tabular}{lccc}
\toprule
\textbf{Trigger} & \textbf{EM (\%)} & \textbf{Timeout (\%)} & \textbf{Cond.\ Acc.\ (\%)} \\
\midrule
Nudge@6 (30\%) & 48.1 & \textbf{1.7} & 49.0 \\
Nudge@8 (40\%) & 51.4 & 2.7 & 52.8 \\
Nudge@10 (50\%) & 52.8 & 4.2 & 55.2 \\
Nudge@12 (60\%) & 52.1 & 8.4 & 56.9 \\
\rowcolor{rowhl}
\textbf{Nudge@14 (70\%)} & \textbf{53.1} & 11.1 & 58.9 \\
Nudge@16 (80\%) & 51.4 & 14.8 & 59.7 \\
No nudge & 49.1 & 24.7 & \textbf{64.6} \\
\midrule
ReAct (no working memory) & 42.0 & 17.8 & 51.1 \\
\bottomrule
\end{tabular}
\end{table}

\paragraph{Is the nudge sufficient without working memory?} A natural objection is that SLEUTH's gains come from the nudge alone, not the structured state. We test this by applying the identical commitment trigger to ReAct (same message at turn 14, no working memory). ReAct+nudge achieves 42.2\% ($+$0.2 vs.\ vanilla ReAct)---effectively unchanged---while SLEUTH+nudge achieves 53.1\% ($+$4.0 over no-nudge SLEUTH). The nudge alone provides no benefit without organized evidence to synthesize from; SLEUTH's structured facts and ranked hypotheses provide the substrate that makes commitment effective. The working memory is the necessary condition---the nudge merely exploits it (full analysis in Appendix~\ref{app:nudge-isolation}).

\subsection{Cross-Model Generalization}
\label{sec:cross-model}

To test whether SLEUTH's advantage reflects the structured protocol rather than model-specific behavior, we replicate key experiments with GLM-5, a model from a different family (Table~\ref{tab:crossmodel}). SLEUTH uses runtime enforcement to ensure protocol adherence on GLM-5 (analyzed in \S\ref{sec:adherence}).

\begin{table}[t]
\caption{Cross-model generalization with GLM-5 (EM, \%). SLEUTH uses runtime working memory enforcement (\S\ref{sec:adherence}). Best result in \textbf{bold} per dataset.}
\label{tab:crossmodel}
\centering
\small
\begin{tabular}{lccccc}
\toprule
\textbf{Method} & \textbf{HotpotQA} & \textbf{2-hop} & \textbf{3-hop} & \textbf{4-hop} & \textbf{2Wiki} \\
\midrule
ReAct & 64.6 & 55.4 & 49.7 & 37.8 & 69.6 \\
ReAct-CoT & 64.3 & 55.6 & 48.8 & 37.0 & 71.2 \\
Self-Ask & 63.9 & 53.8 & 43.9 & 34.1 & 68.4 \\
IRCoT & 56.0 & 43.6 & 36.2 & 23.5 & 59.6 \\
Reflexion & 60.1  & 46.2 & 38.2 & 23.7 & 67.0 \\
\midrule
\rowcolor{rowhl}
\textbf{SLEUTH} & \textbf{65.6} & \textbf{61.5} & \textbf{54.3} & \textbf{48.9} & \textbf{72.3} \\
\midrule
\rowcolor{rowhl}
$\Delta$ vs.\ best baseline & +1.0 & +5.9 & +4.6 & +11.1 & +1.1 \\
\bottomrule
\end{tabular}
\end{table}

SLEUTH with GLM-5 follows the same pattern as with Sonnet: gains grow with difficulty, from +1.0 on HotpotQA to +11.1 on 4-hop. The non-linear jump tracks timeout dynamics---without structured memory, GLM-5's timeout rate climbs from 17\% (2-hop) to 35\% (4-hop), while SLEUTH holds it below 12\% throughout. The modest HotpotQA gain is itself informative: GLM-5's ReAct agent averages 4.3 searches per question (vs.\ 3.0 for Sonnet), brute-forcing easy problems through thoroughness where structure is not yet needed. Notably, Reflexion underperforms ReAct on every multi-hop split, indicating that post-episode reflection without structured within-episode state is insufficient as chains lengthen.

\subsection{Protocol Adherence}
\label{sec:adherence}

The component ablation (\S\ref{sec:scaling}) shows that the \emph{content} of the working memory matters; here we ask whether \emph{consistent maintenance} is necessary. We compare GLM-5 under two configurations: (1)~the general SLEUTH prompt (identical to Sonnet's) with no additional enforcement, and (2)~a model-adapted prompt with explicit structural delimiters and a runtime check that rejects malformed responses. Together, these ensure near-complete protocol adherence (${\sim}$99\% of turns).

\begin{table}[t]
\caption{Effect of protocol adherence on GLM-5 (EM, \%). Without enforcement, the model follows the working memory protocol on only ${\sim}$15\% of turns. With prompt adaptation and runtime validation, adherence rises to ${\sim}$99\%. The benefit scales with difficulty.}
\label{tab:adherence}
\centering
\small
\begin{tabular}{lcccc}
\toprule
\textbf{Enforcement} & \textbf{HotpotQA} & \textbf{2-hop} & \textbf{3-hop} & \textbf{4-hop} \\
\midrule
General prompt (${\sim}$15\% adherence) & 64.4 & 52.6 & 42.6 & 29.9 \\
\rowcolor{rowhl}
Adapted + enforced (${\sim}$99\%) & \textbf{65.6} & \textbf{61.5} & \textbf{54.3} & \textbf{48.9} \\
\midrule
\rowcolor{rowhl}
$\Delta$ & +1.1 & +8.9 & +11.7 & +19.0 \\
\bottomrule
\end{tabular}
\end{table}

Given the same prompt as Sonnet, GLM-5 maintains the protocol on only ${\sim}$15\% of turns, drifting into unstructured reasoning for the rest; prompt adaptation and the runtime check raise this to ${\sim}$99\%. The payoff scales steeply with difficulty---+8.9 on 2-hop, +11.7 on 3-hop, +19.0 on 4-hop. This provides the clearest evidence that \emph{how} an agent organizes its reasoning, not raw model capability, is the active ingredient: a fixed model gains up to 19 points on the hardest problems purely from being held to the structure it was already told to maintain. The working memory must be live and continuously updated---merely seeing the prompt is not enough.

\paragraph{Error analysis.}
\label{sec:error-analysis}
We decompose SLEUTH's remaining errors on 4-hop MuSiQue into three categories: \emph{reasoning errors} (32.1\%)---incorrect inferences from relevant evidence; \emph{timeouts} (11.1\%)---budget exhaustion without committing; and \emph{verbose answers} (4.4\%)---the gold answer is present but includes extraneous context. The primary bottleneck is \emph{inference quality over correctly-retrieved evidence}, not retrieval or state management. In head-to-head comparison, SLEUTH wins 59 questions that ReAct misses while losing only 14, a net advantage of 45 (full decomposition in Appendix~\ref{sec:error-appendix}).

\paragraph{Retrieval sensitivity.}
\label{sec:retrieval}
Our default uses top-$k{=}3$ paragraphs per search. To test whether SLEUTH's advantage depends on this setting, we vary $k \in \{1, 3, 5\}$ on 4-hop MuSiQue. The advantage holds across all three: +10.1 points at top-$k{=}1$, +11.1 at the default $k{=}3$, and +10.9 at top-$k{=}5$ (Appendix~\ref{app:robustness}). The mechanism cuts both ways: under sparse retrieval, fact-tracking preserves a relevant paragraph that might otherwise be buried under later turns, while under noisy retrieval, separating confirmed facts from unverified content helps the agent ignore irrelevant paragraphs.

\paragraph{Two failure modes, two mechanisms.} Our results trace SLEUTH's advantage to addressing two distinct failure modes of implicit state management, each dominant at a different difficulty. \emph{Information loss}---confirmed evidence getting buried under accumulating context---is addressed by structured fact-tracking and accounts for most of the gain on 2- and 3-hop problems, where agents find the evidence but lose it. \emph{Information paralysis}---knowing the answer but failing to commit under bounded computation---is addressed by the commitment mechanism and accounts for most of the gain on 4-hop, where timeouts dominate (cf.\ the 24.7\% no-nudge timeout rate, \S\ref{sec:evidence-sufficiency}). Standard agents suffer both; SLEUTH's structure is what makes each remedy possible---the same organized state that prevents loss is also the substrate the commitment trigger exploits (Appendix~\ref{app:nudge-isolation}).

\section{Discussion}
\label{sec:discussion}

\paragraph{Epistemic state management as a design principle.} SLEUTH achieves substantial improvements through structured state management alone---no additional training, parameters, or trajectory search. This suggests that as agents tackle increasingly complex tasks, the bottleneck shifts from \emph{reasoning capability} to \emph{reasoning organization}~\citep{besta2025demystifying}. The cross-model results make this strikingly concrete: on 4-hop problems, GLM-5 with enforced structure (48.9\%) \emph{outperforms} the stronger Sonnet running unstructured ReAct (42.0\%)---organization on a weaker model beats raw capability without it. Because the working memory is output in natural language on every turn, SLEUTH's reasoning is also inherently more interpretable than that of black-box agents: each step's epistemic state is externalized, enabling human oversight and debugging. As instruction-following improves, we expect structured protocols to become effective without runtime enforcement, extending the benefit to any agentic workflow where state accumulates.

\paragraph{The evidence sufficiency problem.} Our nudge ablation identifies a general tension for agents with rich internal state: the richer the uncertainty model, the harder it becomes to commit under bounded computation. This ``certainty trap'' is not unique to SLEUTH---it arises in any agent framework that makes evidence quality explicit, from autonomous research agents~\citep{nakano2022webgpt} to diagnostic systems. We address it here via a simple time-based nudge; principled alternatives may draw on optimal stopping theory, conformal or coverage-calibrated stopping rules that offer formal guarantees, learned value functions over epistemic states~\citep{chen2023fireact}, or RL-trained stop/continue policies~\citep{jin2025searchr1}. A natural common substrate is SLEUTH's structured state features---fact count, hypothesis confidence, question resolution rate---over which any of these criteria could be calibrated, connecting language agent design to classical decision theory.

\paragraph{Limitations.} SLEUTH's working memory adds token overhead, though it is smaller on the harder problems where the framework helps most (1.33$\times$ the API cost of ReAct on 4-hop vs.\ 1.50$\times$ on HotpotQA; Appendix~\ref{app:tokens}) and benefits most from budgets allowing at least ${\sim}2.5\times$ the reasoning depth, while remaining advantageous even at tighter budgets (Appendix~\ref{app:robustness}). The commitment threshold (70\% of budget) was tuned on our benchmarks and may need re-tuning for very different task structures, and the monotonic fact policy can entrench early reasoning errors when a confirmed fact is contextually misleading (Appendix~\ref{app:discussion}). Finally, our closed-context setting does not capture contradictory or time-sensitive open-domain sources.

\section{Conclusion}

We introduce SLEUTH, a structured working memory framework that decomposes an agent's epistemic state into confirmed facts, ranked hypotheses, and prioritized open questions. SLEUTH's advantage scales with reasoning difficulty, and our analysis isolates two failure modes of implicit state management---information loss and evidence-sufficiency paralysis---each addressed by a distinct component: structured fact-tracking prevents the first, while a commitment mechanism resolves the second, but only when it has organized state to exploit. The central lesson is that \emph{how} an agent organizes its reasoning can matter more than \emph{how well} it reasons: a weaker model held to the structure outperforms a stronger one without it. Because these gains require no training, search, or auxiliary infrastructure---only prompting---epistemic state management emerges as a design dimension for language agents complementary to scaling model size or test-time compute.

%

\section*{Reproducibility Statement}

SLEUTH is implemented entirely through prompting---no fine-tuning, custom architectures, or additional training is required. Complete system prompts for SLEUTH and all baselines and ablations appear in Appendix~\ref{app:prompts}, and the retrieval implementation (lexical/sparse scoring over the per-question paragraph collection) is detailed in Appendix~\ref{app:discussion}. All experiments use publicly available datasets (HotpotQA, MuSiQue, 2WikiMultiHopQA) from the HuggingFace Hub, with statistics and splits in Appendix~\ref{app:datasets}. Unless otherwise noted, all runs use a 20-turn budget, top-$k{=}3$ retrieval, the EM normalization described in \S\ref{sec:experiments}, and greedy decoding (temperature 0); GLM-5 additionally uses the runtime enforcement specified in \S\ref{sec:adherence} and Appendix~\ref{app:discussion}. We fix random seeds for dataset loading and report the model versions used (Claude Sonnet 4.6, GLM-5). As with any hosted model, exact outputs may shift across provider-side model updates; we therefore document all prompts, configurations, and decoding settings needed to reproduce our protocol independent of a specific endpoint.

\bibliography{nliu}
\bibliographystyle{iclr2026_conference}

\newpage
\appendix

\section{Dataset Statistics}
\label{app:datasets}

Table~\ref{tab:datasets} summarizes the benchmarks used in our evaluation. All datasets use the distractor setting, where gold supporting paragraphs are mixed with distractors to create a realistic retrieval challenge. Question and answer lengths are measured in tokens. MuSiQue's controlled hop structure enables our difficulty-scaling analysis (\S\ref{sec:scaling}).

\begin{table}[h]
\caption{Dataset statistics. All use the distractor setting with gold + distractor paragraphs per question.}
\label{tab:datasets}
\centering
\small
\begin{tabular}{lccccc}
\toprule
\textbf{Dataset} & \textbf{$N$} & \textbf{Hops} & \textbf{Avg.\ Q len} & \textbf{Avg.\ A len} & \textbf{Paragraphs} \\
\midrule
HotpotQA & 7{,}405 & 2 & 15.7 & 2.5 & 10 \\
MuSiQue 2-hop & 1{,}252 & 2 & 13.9 & 2.8 & 20 \\
MuSiQue 3-hop & 760 & 3 & 20.9 & 2.9 & 20 \\
MuSiQue 4-hop & 405 & 4 & 25.8 & 2.6 & 20 \\
2WikiMultiHop & 12{,}576 & 2 & 12.0 & 2.4 & 10 \\
\bottomrule
\end{tabular}
\end{table}

\section{F1 Scores}
\label{app:f1}

Table~\ref{tab:f1} reports token-level F1 scores corresponding to the main results in Table~\ref{tab:main}. SLEUTH attains the highest F1 on every benchmark, and its lead widens with difficulty (e.g., 62.2 vs.\ 50.3 for ReAct on 4-hop, a +11.9-point gap that mirrors the EM result). The component ablations track their EM ordering, confirming that the F1 picture reflects genuinely more correct answers rather than merely longer or shorter outputs; the source-faithful extraction rule, which copies answer phrasing verbatim from confirmed facts, keeps SLEUTH's exact-match rate close to its token-overlap rate.

\begin{table}[h]
\caption{F1 scores (\%) across five multi-hop reasoning benchmarks (Claude Sonnet 4.6).}
\label{tab:f1}
\centering
\small
\begin{tabular}{lccccc}
\toprule
\textbf{Method} & \textbf{HotpotQA} & \textbf{2-hop} & \textbf{3-hop} & \textbf{4-hop} & \textbf{2Wiki} \\
\midrule
ReAct & 76.9 & 70.2 & 65.6 & 50.3 & 82.9 \\
ReAct-CoT & 75.1 & 68.9 & 66.0 & 51.2 & 82.4 \\
Self-Ask & 76.0 & 67.7 & 63.6 & 47.4 & 82.1 \\
IRCoT & 74.1 & 68.3 & 64.7 & 48.3 & 81.8 \\
Reflexion & 77.8 & 68.1 & 66.7 & 50.4 & 83.0 \\
\midrule
Notes & 75.2 & 70.0 & 67.8 & 50.8 & 82.2 \\
Facts-Only & 78.2 & 73.6 & 71.7 & 58.8 & 83.5 \\
Facts+Q. & 76.9 & 73.1 & 71.0 & 56.7 & 83.0 \\
Facts+Hyp. & 78.2 & 73.1 & 71.7 & 56.6 & 83.6 \\
\midrule
\rowcolor{rowhl}
\textbf{SLEUTH} & \textbf{80.6} & \textbf{76.8} & \textbf{73.6} & \textbf{62.2} & \textbf{84.5} \\
\bottomrule
\end{tabular}
\end{table}

\section{Token Efficiency}
\label{app:tokens}

Table~\ref{tab:tokens} reports average token usage per question. SLEUTH's working memory output increases token usage modestly on easy tasks (42\% on HotpotQA) and minimally on hard tasks (29\% on 4-hop), while using comparable or fewer tool calls.

\begin{table}[h]
\caption{Average tokens and tool calls per question (Claude Sonnet 4.6). Cost column shows relative API cost vs.\ ReAct (price-weighted, since output tokens are billed higher than input). SLEUTH's raw token overhead shrinks from 42\% on HotpotQA to 29\% on 4-hop (from the Tokens columns) as all agents use more tokens on harder problems; the price-weighted Cost is correspondingly 1.50$\times$ and 1.33$\times$.}
\label{tab:tokens}
\centering
\small
\begin{tabular}{lrrrrrr}
\toprule
& \multicolumn{3}{c}{\textbf{HotpotQA}} & \multicolumn{3}{c}{\textbf{4-hop}} \\
\cmidrule(lr){2-4} \cmidrule(lr){5-7}
\textbf{Method} & \textbf{Tokens} & \textbf{Turns} & \textbf{Cost} & \textbf{Tokens} & \textbf{Turns} & \textbf{Cost} \\
\midrule
ReAct & 6,049 & 3.0 & 1.0$\times$ & 50,786 & 10.3 & 1.0$\times$ \\
ReAct-CoT & 6,268 & 3.0 & 1.07$\times$ & 51,347 & 9.7 & 1.04$\times$ \\
Self-Ask & 6,352 & 3.1 & 1.05$\times$ & 50,976 & 10.9 & 1.00$\times$ \\
IRCoT & 6,521 & 3.0 & 1.13$\times$ & 55,952 & 10.5 & 1.17$\times$ \\
Reflexion & 20,519 & 12.1 & 3.39$\times$ & 100,961 & 32.6 & 2.00$\times$ \\
Notes & 6,332 & 2.9 & 1.11$\times$ & 61,604 & 9.8 & 1.29$\times$ \\
Facts-Only & 6,879 & 3.1 & 1.17$\times$ & 52,751 & 10.6 & 1.06$\times$ \\
Facts+Hyp. & 6,710 & 3.0 & 1.16$\times$ & 56,272 & 10.7 & 1.12$\times$ \\
\rowcolor{rowhl}
\textbf{SLEUTH} & 8,565 & 3.0 & 1.50$\times$ & 65,751 & 10.7 & 1.33$\times$ \\
\bottomrule
\end{tabular}
\end{table}

\section{The Commitment Mechanism: Supporting Analyses}
\label{app:commitment}

This appendix collects the analyses underpinning the evidence sufficiency problem and the nudge mechanism (\S\ref{sec:evidence-sufficiency}): the timeout/conditional-accuracy decomposition that defines the problem, the commitment-timing sweep, accuracy-at-turn convergence, the nudge-isolation control showing structure is necessary, and a comparison against alternative stopping strategies.

\subsection{Timeout Decomposition}
\label{app:timeout}

Table~\ref{tab:timeout} decomposes EM into conditional accuracy (accuracy among answered questions) and timeout rate (fraction of questions where the agent exhausts its 20-turn budget). A key insight: on 4-hop, SLEUTH achieves both higher conditional accuracy (58.9\% vs.\ 51.1\%) \emph{and} lower timeout rate (11.1\% vs.\ 17.8\%). The 11.1-point EM gap reflects improvements on both dimensions: the structured working memory enables better reasoning, while the commitment mechanism ensures that reasoning translates into answers.

\begin{table}[h]
\caption{Timeout decomposition (Claude Sonnet 4.6). Cond.\ Acc.\ = EM among answered questions; Timeout = fraction of budget-exhausting episodes.}
\label{tab:timeout}
\centering
\small
\begin{tabular}{llccc}
\toprule
\textbf{Dataset} & \textbf{Method} & \textbf{EM (\%)} & \textbf{Timeout (\%)} & \textbf{Cond.\ Acc.\ (\%)} \\
\midrule
\multirow{2}{*}{HotpotQA} & ReAct & 60.2 & 0.3 & 60.4 \\
& SLEUTH & 66.1 & 0.2 & 66.2 \\
\midrule
\multirow{2}{*}{2-hop} & ReAct & 54.6 & 2.8 & 56.2 \\
& SLEUTH & 66.5 & 2.2 & 67.8 \\
\midrule
\multirow{2}{*}{3-hop} & ReAct & 52.0 & 3.8 & 54.0 \\
& SLEUTH & 63.7 & 4.1 & 66.0 \\
\midrule
\multirow{2}{*}{4-hop} & ReAct & 42.0 & 17.8 & 51.1 \\
& SLEUTH & 53.1 & 11.1 & 58.9 \\
\bottomrule
\end{tabular}
\end{table}

\subsection{Nudge Ablation Visualization}
\label{app:nudge-fig}

\begin{figure}[h]
\centering
\includegraphics[width=0.85\linewidth]{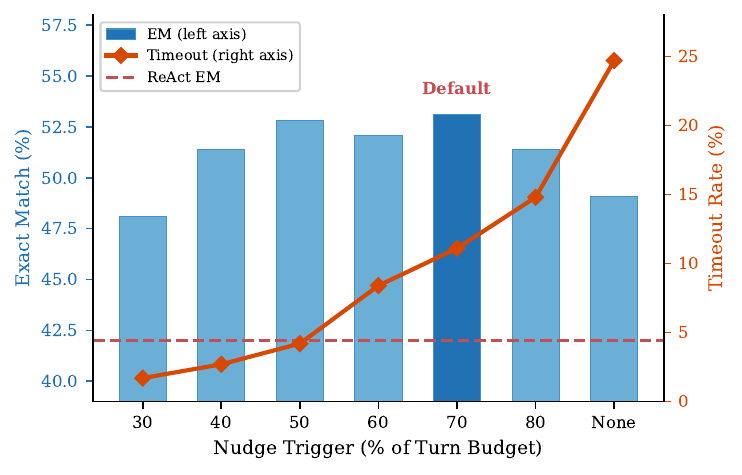}
\caption{Commitment timing ablation on 4-hop MuSiQue. Blue bars show EM (left axis); orange line shows timeout rate (right axis). Performance plateaus in the 50--70\% range, with all configurations outperforming ReAct (dashed red line). We use 70\% as the default.}
\label{fig:nudge}
\end{figure}

Figure~\ref{fig:nudge} visualizes the exploration-exploitation tradeoff from Table~\ref{tab:nudge}. EM plateaus in the 50--70\% range while timeout rate increases monotonically with later nudges.

\paragraph{Contrast with easy questions.} On HotpotQA (2-hop) with GLM-5, the same nudge sweep yields a total spread of only 0.55pp (EM ranges from 64.9 to 65.4 across thresholds $\alpha \in [0.2, 0.8]$), confirming that the commitment mechanism's value is concentrated on hard problems where the evidence sufficiency problem is acute. On easy questions, most episodes terminate naturally before any nudge fires.

\subsection{Convergence Analysis}
\label{app:convergence}

\begin{figure}[h]
\centering
\includegraphics[width=\linewidth]{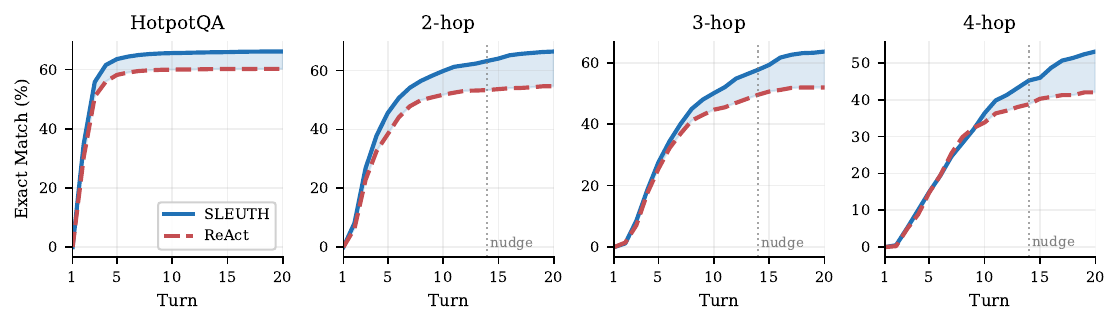}
\caption{Accuracy at turn $t$ across four benchmarks. On HotpotQA, both methods converge by turn 5, but SLEUTH reaches a higher final accuracy. On multi-hop problems, the two methods track closely during exploration; SLEUTH pulls ahead after the commitment mechanism fires (dotted line at turn 14), converting accumulated evidence into answers while ReAct plateaus due to timeouts.}
\label{fig:convergence}
\end{figure}

Figure~\ref{fig:convergence} shows \emph{accuracy at turn $t$}: the EM that would result if all agents were forced to stop at turn $t$ (episodes not yet terminated count as incorrect). This reveals how quickly each method converts evidence into correct answers.

Two patterns emerge. On HotpotQA (easy 2-hop), both methods converge by turn 5, but SLEUTH reaches a higher plateau---the working memory enables more accurate answers even without the nudge, because the structured decomposition prevents redundant searches and aids inference. On harder problems (3- and 4-hop), the curves are nearly identical through turn 12--14, then diverge sharply. This confirms that SLEUTH's advantage on hard problems comes primarily from the commitment mechanism converting potential timeouts into answers, while its advantage on easy problems comes from better reasoning quality via structured state tracking.

\subsection{Nudge Isolation}
\label{app:nudge-isolation}

A natural question arises from the convergence analysis (\S\ref{app:convergence}): if SLEUTH and ReAct perform similarly before turn 14, does the nudge alone explain SLEUTH's advantage? We test this by applying the same commitment trigger to ReAct---identical nudge message at turn 14, without any working memory structure.

\begin{table}[h]
\caption{Nudge isolation on 4-hop MuSiQue (EM, \%). The commitment trigger has no effect on ReAct but helps SLEUTH by +4.0pp, demonstrating that structured working memory is necessary for effective commitment.}
\label{tab:nudge-isolation}
\centering
\small
\begin{tabular}{lcc}
\toprule
\textbf{Method} & \textbf{EM (\%)} & \textbf{Effect of nudge} \\
\midrule
ReAct & 42.0 & --- \\
ReAct + nudge@14 & 42.2 & +0.2 \\
\midrule
SLEUTH (no nudge) & 49.1 & --- \\
\rowcolor{rowhl}
SLEUTH + nudge@14 & \textbf{53.1} & +4.0 \\
\bottomrule
\end{tabular}
\end{table}

The nudge has no effect on ReAct ($+$0.2pp, within noise): forced to commit at turn 14, ReAct's performance is unchanged because its conversation history is a flat sequence of searches and thoughts with no organized state to synthesize from---the nudge simply triggers whatever the agent would have guessed anyway. In contrast, SLEUTH gains +4.0pp from the nudge because its working memory provides structured facts and ranked hypotheses ready for synthesis. This result establishes that the working memory is the \emph{necessary condition} for effective commitment. The nudge is not independently beneficial---it is a mechanism that \emph{exploits} the organized epistemic state that SLEUTH maintains. The same argument applies to other baselines (IRCoT, Self-Ask): none maintain persistent organized state across turns, so a commitment trigger would find no structured evidence to synthesize from.

\paragraph{Connection to exploration-exploitation theory.} The nudge isolation result can be understood through the lens of the exploration-exploitation tradeoff in sequential decision-making. In bandit settings, forced exploitation (greedy action selection) is harmful when the agent's estimate of reward is noisy---it exploits an unreliable signal. The analogous situation here: ReAct's implicit state after 14 turns is a noisy, unorganized signal---the agent has encountered the relevant facts but has no structured representation of what they collectively imply. Forcing commitment exploits noise.

SLEUTH's working memory acts as a \emph{sufficient statistic} for the evidence gathered: the confirmed facts and ranked hypotheses compress the full trajectory into a decision-relevant representation. When the nudge fires, the agent synthesizes from this compressed state rather than the raw conversation history. The quality of the commitment depends on the quality of this compression---which is precisely what structured working memory provides.

\paragraph{Implications for agent design.} This decomposition---working memory as state compression, nudge as exploitation trigger---suggests a general design principle for bounded-computation agents: before adding commitment mechanisms, ensure the agent maintains state in a form suitable for commitment. Concretely, any agent framework that (a) maintains structured intermediate state and (b) provides a principled exploitation trigger should achieve the same synergy. The working memory need not follow SLEUTH's exact tripartite format; what matters is that it separates signal (confirmed evidence) from noise (raw observations) and rankings (hypotheses) from uncertainties (open questions), enabling a single synthesis step to produce a coherent answer.

\subsection{Alternative Stopping Strategies}
\label{app:variants}

The commitment mechanism (nudge at $\alpha{=}0.7$) is one solution to the evidence sufficiency problem. We evaluated four alternative strategies on 4-hop MuSiQue, each embodying a different principled answer to ``when should an agent commit?''

\begin{table}[h]
\caption{Alternative stopping strategies on 4-hop MuSiQue (Claude Sonnet 4.6). All use the same working memory structure; only the commitment strategy differs.}
\label{tab:variants}
\centering
\small
\begin{tabular}{llccc}
\toprule
\textbf{Variant} & \textbf{Strategy} & \textbf{EM (\%)} & \textbf{Timeout (\%)} & \textbf{Cond.\ Acc.\ (\%)} \\
\midrule
No commitment & Open-ended (baseline) & 49.1 & 24.7 & \textbf{64.6} \\
\midrule
Structural completion & Commit when all sub-Qs resolved & 47.2 & 18.8 & 57.8 \\
Forward-only & Linear progression, no re-opening & 45.7 & 16.8 & 54.6 \\
Contrastive elimination & Search to distinguish hypotheses & 47.2 & 19.0 & 57.9 \\
Late verification & Free search + chain check at end & 42.5 & 29.9 & 59.5 \\
\midrule
\rowcolor{rowhl}
\textbf{Nudge ($\alpha{=}0.7$)} & \textbf{Time-based commitment trigger} & \textbf{53.1} & \textbf{11.1} & 58.9 \\
\bottomrule
\end{tabular}
\end{table}

The time-based nudge outperforms all structural alternatives by a wide margin (+5.9 to +10.6 EM). We analyze why each structural strategy fails:

\paragraph{Why structural completion fails.} This strategy requires the agent to decompose the question into sub-questions upfront and commit when all are marked ``resolved.'' The failure mode is twofold: (1) the agent's initial decomposition may be incomplete or poorly scoped---a 4-hop question might be decomposed into 3 sub-questions, causing premature commitment; and (2) the ``resolved'' criterion is itself ambiguous---the agent may mark a sub-question resolved based on weak evidence to satisfy the structural requirement. Timeout rate drops modestly (18.8\% vs.\ 24.7\%) but conditional accuracy falls 6.8pp, indicating that the commitment criterion triggers at the wrong time.

\paragraph{Why forward-only commitment fails.} By preventing the agent from revisiting earlier conclusions, this strategy eliminates over-verification but also eliminates \emph{beneficial} re-evaluation. On 4-hop chains, early errors propagate: if the agent incorrectly identifies an intermediate entity (e.g., the wrong birth state), the forward-only constraint prevents correction even when contradicting evidence appears later. The lowest conditional accuracy (54.6\%) among the structural strategies confirms that irreversibility is too strong a constraint.

\paragraph{Why contrastive elimination fails.} Searching for evidence that \emph{distinguishes} hypotheses is in principle more efficient than confirmatory search, but in practice the agent struggles to formulate effective contrastive queries. Multi-hop questions often have only one viable hypothesis by mid-trajectory; the contrastive protocol then generates artificial alternatives to ``eliminate,'' wasting turns on searches for non-existent distinctions. The 19.0\% timeout rate reflects this wasted effort.

\paragraph{Why late verification fails.} This strategy separates search (free-form) from verification (chain check before answering). The intent is to impose structure only at decision time, avoiding per-turn overhead. However, without continuous state tracking, the free-form search phase suffers from the same context dilution as ReAct---by the time verification occurs, the agent cannot reliably reconstruct the full chain from its unorganized history. The worst EM (42.5\%) and highest timeout (29.9\%) confirm that structure imposed only at the end cannot compensate for the lack of ongoing state management.

\paragraph{The time-based nudge succeeds} because it does not constrain \emph{how} the agent reasons or \emph{what criteria} define sufficiency---it simply bounds the exploration phase. The agent retains full freedom during the first 70\% of its budget and receives one intervention that triggers synthesis from existing structured state. This separates the commitment \emph{decision} (which the model makes) from the commitment \emph{trigger} (which the mechanism provides), exploiting rather than overriding the model's reasoning.

\section{Working Memory Dynamics}
\label{app:wm-dynamics}

\begin{figure}[h]
\centering
\includegraphics[width=\linewidth]{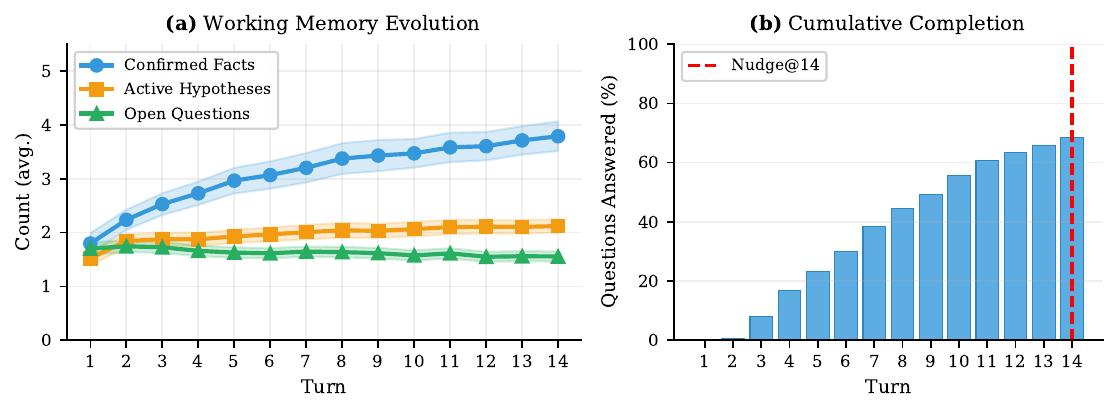}
\caption{Working memory dynamics on 4-hop MuSiQue. (a) Average component counts per turn (95\% CIs, fixed cohort of episodes ${\geq}14$ turns). Facts accumulate monotonically while open questions decrease, demonstrating progressive uncertainty resolution. (b) Cumulative fraction of episodes that terminate naturally (agent calls \texttt{finish}) by each turn. About two-thirds of episodes complete before the nudge fires (dashed red, turn~14); the nudge intervenes only in the remaining ${\sim}$34\% that would otherwise risk timeout.}
\label{fig:wm-evolution}
\end{figure}

Figure~\ref{fig:wm-evolution} visualizes how the working memory evolves across turns on 4-hop MuSiQue ($n{=}405$), restricted to episodes lasting ${\geq}14$ turns to avoid survivorship bias. Panel (a) shows evidence monotonicity in practice: confirmed facts accumulate steadily across the trajectory, reflecting continuous evidence extraction. Active hypotheses remain stable (${\sim}2$ throughout), reflecting the agent's ability to maintain and rank competing explanations without proliferation. Open questions decrease gradually, reflecting progressive uncertainty resolution. Panel (b) shows that ${\sim}$66\% of episodes terminate naturally before the nudge at turn~14 (the agent finds sufficient evidence and calls \texttt{finish} without intervention). The commitment mechanism fires only in the remaining ${\sim}$34\% of episodes---the fixed cohort plotted in panel (a)---precisely those that would otherwise risk timeout.

\section{Robustness Analysis}
\label{app:robustness}

We test SLEUTH's sensitivity to two key hyperparameters: turn budget and retrieval breadth.

\paragraph{Turn budget.} Table~\ref{tab:budget} varies the maximum number of turns on 4-hop MuSiQue. For each budget, we sweep the nudge timing and report the best configuration (turn 6, 10, and 14 for budgets 10, 15, and 20 respectively---all near $\alpha{=}0.7$).

\begin{table}[h]
\caption{Effect of turn budget on 4-hop MuSiQue (EM, \%). For each budget, SLEUTH uses the best nudge timing from a sweep ($\alpha \approx 0.7$). SLEUTH dominates ReAct at every budget level, with a ${\sim}$9--11pp advantage.}
\label{tab:budget}
\centering
\small
\begin{tabular}{lccc}
\toprule
\textbf{Budget} & \textbf{SLEUTH} & \textbf{ReAct} & \textbf{$\Delta$} \\
\midrule
10 turns & \textbf{42.7} & 34.1 & +8.6 \\
15 turns & \textbf{52.1} & 40.7 & +11.4 \\
\rowcolor{rowhl}
20 turns & \textbf{53.1} & 42.0 & +11.1 \\
\bottomrule
\end{tabular}
\end{table}

SLEUTH's advantage is robust across budgets (+8.6 to +11.4pp), demonstrating that the structured working memory provides consistent benefit regardless of computational constraints. Notably, SLEUTH saturates at budget=15 (52.1\%), nearly matching its budget=20 performance (53.1\%)---the commitment mechanism at $\alpha{=}0.7$ effectively resolves all questions that are resolvable within the evidence space. ReAct continues to improve with additional budget (34.1$\to$40.7$\to$42.0) but never closes the gap, confirming that SLEUTH's advantage comes from \emph{organizing} evidence rather than merely having more time. This confirms that the ${\sim}$2.5$\times$ hops guideline noted in \S\ref{sec:discussion} is not a hard constraint but a soft one: SLEUTH benefits at any budget where at least one full reasoning chain can be completed.

\paragraph{Retrieval breadth.} Table~\ref{tab:topk} varies the number of paragraphs returned per search query on 4-hop MuSiQue. SLEUTH's advantage is substantial under sparse (top-$k{=}1$: +10.1pp), default (top-$k{=}3$: +11.1pp), and noisy (top-$k{=}5$: +10.9pp) retrieval. With sparse retrieval, the working memory's fact-tracking prevents information loss when relevant paragraphs are found only once across multiple searches. With noisy retrieval, the structured fact/hypothesis separation helps the agent distinguish signal from irrelevant context. The consistent ${\sim}$10--11pp advantage across all retrieval settings confirms that SLEUTH's benefit is robust to retrieval quality.

\begin{table}[h]
\caption{Effect of retrieval breadth (top-$k$) on 4-hop MuSiQue (EM, \%). SLEUTH's advantage grows under both sparse and noisy retrieval conditions.}
\label{tab:topk}
\centering
\small
\begin{tabular}{lccc}
\toprule
\textbf{top-$k$} & \textbf{SLEUTH} & \textbf{ReAct} & \textbf{$\Delta$} \\
\midrule
1 & 46.9 & 36.8 & +10.1 \\
\rowcolor{rowhl}
3 (default) & \textbf{53.1} & 42.0 & \textbf{+11.1} \\
5 & 53.6 & 42.7 & +10.9 \\
\bottomrule
\end{tabular}
\end{table}

\section{Statistical Significance}
\label{app:significance}

We report bootstrap confidence intervals (10{,}000 resamples, paired by question) for the key comparison: SLEUTH vs.\ ReAct with Claude Sonnet 4.6. All differences are significant at $p < 0.001$. Note that confidence intervals widen with difficulty (from $\pm$0.7 on HotpotQA to $\pm$4.0 on 4-hop) due to decreasing sample size ($N{=}7405 \to 405$), yet even the lower bound of the 4-hop CI (+7.2) exceeds the HotpotQA point estimate, confirming that the scaling-with-difficulty finding is robust.

\begin{table}[h]
\caption{Bootstrap 95\% confidence intervals for SLEUTH $-$ ReAct EM difference.}
\label{tab:bootstrap}
\centering
\small
\begin{tabular}{lccc}
\toprule
\textbf{Dataset} & \textbf{Mean $\Delta$} & \textbf{95\% CI} & \textbf{$p$-value} \\
\midrule
HotpotQA & +5.8 & [+5.2, +6.6] & $<$0.0001 \\
MuSiQue 2-hop & +11.8 & [+9.7, +14.0] & $<$0.0001 \\
MuSiQue 3-hop & +11.7 & [+8.9, +14.5] & $<$0.0001 \\
MuSiQue 4-hop & +11.1 & [+7.2, +15.3] & $<$0.0001 \\
\bottomrule
\end{tabular}
\end{table}

\section{Head-to-Head Analysis}
\label{app:headtohead}

Table~\ref{tab:headtohead} shows per-question win/loss counts between SLEUTH and ReAct. SLEUTH's advantage is broad-based: it wins approximately 4--6$\times$ more questions than it loses on every dataset. The ``both wrong'' category is largest on 4-hop (176/405 = 43\%), indicating substantial headroom---these are questions where better retrieval or inference could help both methods, and where complementary approaches (e.g., deliberative verification) would add value orthogonally to SLEUTH's state management contribution.

\begin{table}[h]
\caption{Per-question outcomes: SLEUTH vs.\ ReAct (Claude Sonnet 4.6). ``SLEUTH wins'' = SLEUTH correct, ReAct wrong; ``ReAct wins'' = vice versa.}
\label{tab:headtohead}
\centering
\small
\begin{tabular}{lccccr}
\toprule
\textbf{Dataset} & \textbf{SLEUTH wins} & \textbf{ReAct wins} & \textbf{Both correct} & \textbf{Both wrong} & \textbf{Ratio} \\
\midrule
HotpotQA & 571 & 138 & 4{,}321 & 2{,}375 & 4.1$\times$ \\
2-hop & 179 & 31 & 653 & 389 & 5.8$\times$ \\
3-hop & 110 & 21 & 374 & 255 & 5.2$\times$ \\
4-hop & 59 & 14 & 156 & 176 & 4.2$\times$ \\
2Wiki & 581 & 163 & 9{,}027 & 2{,}805 & 3.6$\times$ \\
\bottomrule
\end{tabular}
\end{table}

\section{Working Memory Schema Sensitivity}
\label{app:schema}

The ablation block of Table~\ref{tab:main} includes three two-component variants---Facts-Only (F), Facts+Questions (F+Q), and Facts+Hypotheses (F+H)---that each maintain a different subset of SLEUTH's working memory while sharing the identical commitment mechanism (nudge at $\alpha{=}0.7$) and source-faithful extraction. This isolates the effect of \emph{which} structural components are maintained, and answers whether SLEUTH's full tripartite schema is over-specified: could a single added component (hypotheses \emph{or} questions, but not both) recover the full gain?

Two observations emerge. First, the three two-component variants cluster tightly---within ${\sim}$2pp of each other on every split (e.g., 47.7--49.9 on 4-hop)---and none approaches full SLEUTH, which leads the best two-component variant by 3.2--4.9 points on HotpotQA and MuSiQue (and a smaller +1.4 on the easier 2Wiki, consistent with structure mattering most as difficulty grows). Second, open questions are \emph{not} independently beneficial when added to facts alone: Facts+Questions is the weakest of the three subsets on \emph{every} split, consistently \emph{below} Facts-Only. This indicates that open questions do not help by themselves---they require ranked hypotheses to anchor what each question is trying to resolve. The full tripartite structure is thus synergistic rather than additive: hypotheses give the agent candidate answers to discriminate between, and open questions turn that discrimination into directed action. Removing either collapses the benefit, confirming that each component earns its place only in the presence of the others.

\section{Error Decomposition}
\label{sec:error-appendix}

Table~\ref{tab:error-decomp} decomposes failures into three categories: \emph{timeouts} (agent exhausts turn budget without answering), \emph{reasoning errors} (agent retrieves relevant evidence but produces an incorrect answer), and \emph{verbose answers} (prediction contains the gold answer but includes extraneous context that causes exact-match failure). The ``Other'' column captures remaining partial-match failures where the gold is \emph{not} contained in the prediction (e.g., wrong entity from a related passage).

\begin{table}[h]
\caption{Error decomposition (\% of total questions) for SLEUTH and ReAct across MuSiQue splits (Claude Sonnet 4.6). SLEUTH reduces timeouts and reasoning errors at every difficulty level while maintaining comparable verbose answer rates.}
\label{tab:error-decomp}
\centering
\small
\begin{tabular}{llccccc}
\toprule
\textbf{Split} & \textbf{Method} & \textbf{EM (\%)} & \textbf{Timeout} & \textbf{Reasoning} & \textbf{Verbose} & \textbf{Other} \\
\midrule
\multirow{2}{*}{2-hop} & SLEUTH & 66.5 & 2.2 & 23.3 & 8.1 & 0.0 \\
 & ReAct & 54.6 & 2.8 & 29.3 & 13.3 & 0.0 \\
\midrule
\multirow{2}{*}{3-hop} & SLEUTH & 63.7 & 4.1 & 25.4 & 7.2 & 0.0 \\
 & ReAct & 52.0 & 3.8 & 33.2 & 11.1 & 0.0 \\
\midrule
\multirow{2}{*}{4-hop} & SLEUTH & 53.1 & 11.1 & 32.1 & 4.4 & 0.0 \\
 & ReAct & 42.0 & 17.8 & 35.8 & 4.4 & 0.0 \\
\bottomrule
\end{tabular}
\end{table}

Three findings emerge. First, SLEUTH's timeout rate scales much more gracefully: 2.2\%$\to$4.1\%$\to$11.1\% versus ReAct's 2.8\%$\to$3.8\%$\to$17.8\%. The commitment mechanism effectively addresses the evidence sufficiency problem---without it, SLEUTH's timeout rate on 4-hop is 24.7\% (\S\ref{sec:nudge-ablation}). Second, SLEUTH achieves lower reasoning error rates at every difficulty level ($-$6.0pp on 2-hop, $-$7.8pp on 3-hop, $-$3.7pp on 4-hop), confirming that the structured working memory helps the model draw correct inferences. On 4-hop, SLEUTH's EM is 11 points higher because its lower timeout rate (11.1\% vs.\ 17.8\%) and lower reasoning error rate combine to produce substantially more correct answers. Third, SLEUTH achieves substantially lower verbose answer rates (4.4--8.1\% vs.\ 4.4--13.3\% for ReAct), as the source-faithful extraction rule successfully reduces paraphrasing errors by anchoring answers to confirmed facts.

This per-split decomposition substantiates the two-failure-mode account in \S\ref{sec:retrieval}: SLEUTH's reasoning-error reduction (information loss prevented by fact-tracking) dominates the gain on 2- and 3-hop, while its timeout reduction (information paralysis prevented by the commitment mechanism) dominates on 4-hop. Among SLEUTH's remaining reasoning errors on 4-hop, the primary bottleneck is \emph{inference quality over correctly-retrieved evidence}---the model draws wrong conclusions from the right facts. This bottleneck is orthogonal to SLEUTH's contribution (state organization) and would benefit from complementary approaches such as deliberative verification~\citep{jiang2025ragstar}, context compression~\citep{li2025brief}, or logic-aware retrieval~\citep{liu2025hoprag}.

\section{System Prompts}
\label{app:prompts}

All methods share identical tool interfaces (\texttt{search}, \texttt{lookup}, \texttt{finish}) and a basic answer format instruction (brief factual answer, no explanation). SLEUTH additionally includes a source-faithful extraction rule that instructs the agent to copy answers verbatim from its confirmed facts---this leverages the structured working memory's sourced citations to eliminate paraphrasing errors. This instruction is architecturally motivated: it references SLEUTH's typed \texttt{[F]} entries with source citations, which only exist within the working memory structure; an equivalent ``use exact source wording'' instruction is already included in the shared format for all agents but lacks a structured target to reference. Below we show the \emph{reasoning protocol} portion of each prompt---the only component that differs between methods.

\begin{promptbox}[Shared Answer Format (appended to all methods)]
The answer in finish() must be ONLY the final answer --- a short entity name, number, date, yes/no, or brief phrase. Do NOT include explanations or reasoning. Use the EXACT wording from your source when possible (e.g., if source says ``twice'', answer ``twice'' not ``2'').
\end{promptbox}

\begin{promptbox}[SLEUTH Source-Faithful Extraction]
The answer in finish() must use the EXACT wording from your source text as recorded in your Confirmed Facts. Do NOT paraphrase, abbreviate, expand, or reformat. Copy the answer VERBATIM from your [F] entries --- the exact characters in your confirmed facts are your answer. For yes/no questions: answer ONLY ``yes'' or ``no''. For entity questions: use the EXACT form as it appears in your source (e.g., if your fact says ``1964 to 1974'', answer ``1964 to 1974'' --- not ``1964-1974'').
\end{promptbox}

\begin{promptbox}[ReAct --- Reasoning Protocol]
You are a research assistant answering multi-hop questions that require finding and connecting information from multiple sources.

Process:
1. Think about what the question is asking and what intermediate information you need.
2. Use search(query) to find relevant paragraphs, or lookup(title) if you know the exact title.
3. After each tool result, reason about what you learned and what you still need to find.
4. Once you have enough evidence to answer confidently, call finish(answer).
\end{promptbox}

\begin{promptbox}[ReAct-CoT --- Reasoning Protocol]
You are a research assistant answering multi-hop questions that require finding and connecting information from multiple sources.

Before EVERY action, you must write a detailed chain-of-thought reasoning that:
1. Summarizes what you know so far from previous tool results.
2. Identifies what information is still missing to answer the question.
3. Explains WHY your next tool call will help fill that gap.

Process:
1. Analyze the question carefully. Identify what sub-facts you need.
2. Write your chain-of-thought reasoning.
3. Use search(query) or lookup(title) to find information.
4. After each tool result, write another chain-of-thought before your next action.
5. Once you have enough evidence, use finish(answer).
\end{promptbox}

\begin{promptbox}[Self-Ask --- Reasoning Protocol]
You are a research assistant answering multi-hop questions by decomposing them into simpler sub-questions and answering each one sequentially.

Format your reasoning as:\\
Sub-question 1: [question]\\
{[}use tools to find the answer{]}\\
Intermediate answer 1: [answer]\\
\\
Sub-question 2: [question based on previous answers]\\
{[}use tools to find the answer{]}\\
Intermediate answer 2: [answer]\\
\\
...continue until you can answer the original question, then call finish(answer).
\end{promptbox}

\begin{promptbox}[IRCoT --- Reasoning Protocol]
You are a research assistant answering multi-hop questions using an interleaved retrieval and reasoning approach.

You MUST strictly alternate between REASONING and RETRIEVAL steps:

{[}REASONING{]} Based on what I know: <summary>. I still need to find: <gap>. I will search for: <plan>.\\
{[}RETRIEVAL{]} <tool call>\\
{[}REASONING{]} From the retrieval I learned: <new info>. Combined with what I knew: <updated understanding>. I still need: <next gap or ``nothing, I can answer''>.\\
{[}RETRIEVAL{]} <tool call or finish>

Continue alternating until you can answer, then call finish(answer).
\end{promptbox}

\begin{promptbox}[Reflexion --- Reasoning Protocol]
You are a research assistant answering multi-hop questions that require finding and connecting information from multiple sources.

Process:
1. Think about what the question is asking and what intermediate information you need.
2. Use search(query) or lookup(title) to find relevant paragraphs.
3. After each tool result, reason about what you learned and what you still need.
4. Once you have enough evidence, call finish(answer).

\emph{On retry after incorrect answer:} You are retrying a question you previously got wrong. Reflection: <what went wrong, what to do differently>. Use this reflection to guide a more careful investigation.
\end{promptbox}

\begin{promptbox}[SLEUTH --- Reasoning Protocol]
You are an investigative research assistant. You maintain a structured working memory to organize your investigation systematically.

Working Memory Protocol --- After EVERY tool call, output your COMPLETE working memory:

Confirmed Facts: Verified information from tool results.
\quad - [F1] <statement> (source: <title>)

Active Hypotheses: Possible answers, ranked by evidence.
\quad - [H1] <hypothesis> | confidence: high/medium/low
\quad\quad\quad | supports: <fact IDs> | contradicts: <fact IDs>

Open Questions: What you still need to find out.
\quad - [Q1] <question> | action: <search terms> | priority: high/medium/low

Investigation Loop:
1. Extract new confirmed facts from tool results.
2. Update hypothesis rankings based on new evidence.
3. Mark resolved questions. Generate new if needed.
4. Execute highest-priority open question's action.
5. When a hypothesis reaches high confidence, call finish() IMMEDIATELY.
\end{promptbox}

\noindent \textbf{Ablation variants.} The structural ablations (Notes, Facts-Only, Facts+Hyp.) use prompts of comparable length to SLEUTH, differing only in which working memory components are included. Notes uses an unstructured scratchpad; Facts-Only maintains only the Confirmed Facts section; Facts+Hyp.\ adds ranked hypotheses but omits Open Questions. All include the same shared answer format and tool instructions above.

\section{Example Trace}
\label{app:trace}

Figure~\ref{fig:trace} shows a comparative trace on a real 4-hop MuSiQue question where SLEUTH succeeds in 11 turns and ReAct times out at 20. Both agents achieve 100\% supporting paragraph recall---demonstrating that the failure is not retrieval but \emph{state management}. The reasoning chain requires: album $\to$ performer $\to$ birth state $\to$ largest city $\to$ race winner.

\begin{figure}[h]
\centering
\fbox{\parbox{0.95\linewidth}{\small
\textbf{Question:} ``Who won the Indy Car Race in the largest populated city of the state where the performer of Mingus Plays Piano is from?''\\
\textbf{Gold answer:} Mario Andretti \quad \textbf{Gold paragraphs:} Mingus Plays Piano, Charles Mingus, Tucson Arizona, Desert Diamond West Valley Phoenix Grand Prix

\vspace{0.3em}
\textbf{SLEUTH (succeeds in 11 turns, 15 tool calls):}\\
Turn 1: Decomposes into [Q1] Who performed Mingus Plays Piano? [Q2] What state are they from? [Q3] Largest city in that state? [Q4] IndyCar race winner there?\\
Turn 2: [F1] ``Mingus Plays Piano'' is by Charles Mingus (source: Mingus Plays Piano). [Q1] resolved.\\
Turn 4: [F2] Charles Mingus born in Nogales, Arizona (source: Charles Mingus). [Q2] resolved.\\
Turn 6: [F3] Largest city in Arizona is Phoenix (source: Tucson, Arizona). [Q3] resolved.\\
Turn 8: Searches ``IndyCar race Phoenix'' $\to$ finds Desert Diamond West Valley Phoenix Grand Prix.\\
Turn 11: [F4] Mario Andretti won (source: Desert Diamond GP). All Qs resolved $\to$ \texttt{finish(``Mario Andretti'')}.

\vspace{0.3em}
\textbf{ReAct (times out at turn 20, 22 tool calls):}\\
Turns 1--3: Correctly finds Charles Mingus performed the album, born in Arizona.\\
Turns 4--8: Searches for ``IndyCar Arizona'' but retrieves irrelevant races. Does not track that Phoenix is the target city.\\
Turns 9--14: Re-searches Charles Mingus biography and ``largest city Arizona'' (already established).\\
Turns 15--20: Repeatedly retrieves Desert Diamond GP article but fails to extract the winner, searching instead for ``Phoenix Grand Prix winner'' and ``IndyCar Phoenix results.'' Times out.
}}
\caption{Real 4-hop trace comparison (MuSiQue ID: 4hop1\_\_802394). Both agents retrieve all 4 gold paragraphs (recall = 1.0). SLEUTH's working memory tracks resolved sub-questions, preventing redundant searches and enabling systematic chain completion. ReAct revisits resolved questions (turns 9--14) and fails to connect its findings into the final answer.}
\label{fig:trace}
\end{figure}

\section{Design Rationale}
\label{app:discussion}

We address several design questions that arise from SLEUTH's architecture.

\paragraph{Retrieval implementation.}
The \texttt{search(query)} tool uses lexical (sparse) scoring over the paragraph collection provided with each question (10 paragraphs for HotpotQA/2Wiki, 20 for MuSiQue): each paragraph is ranked by token overlap with the query (a Jaccard-style overlap over word tokens). The query is the agent's free-form search string; the top-$k$ paragraphs by this score are returned verbatim. The \texttt{lookup(title)} tool returns the full text of a named paragraph if it exists in the collection. Both tools are identical across all methods---no method-specific retrieval tuning is applied. Supporting paragraph recall (fraction of gold paragraphs retrieved at least once) exceeds 90\% for all methods on 4-hop MuSiQue, confirming that retrieval quality is not the bottleneck; the difference lies in how agents \emph{organize and act on} retrieved evidence.

\paragraph{Enforcement fairness and budget accounting.}
A natural concern is whether the protocol adherence mechanism (\S\ref{sec:adherence}) introduces an asymmetric advantage. Enforcement consists of two components: (1)~a model-adapted prompt with explicit structural delimiters that raises adherence from ${\sim}$15\% to ${\sim}$90\%, and (2)~a runtime check that rejects responses missing proper notation and re-prompts the agent, raising adherence to ${\sim}$99\%. Rejected turns do \emph{not} count against the 20-turn budget; the mechanism allows up to 5 reminders per episode. This combined enforcement is only applied in the cross-model experiments with GLM-5 (Table~\ref{tab:crossmodel} and Table~\ref{tab:adherence}). The primary Sonnet results in Table~\ref{tab:main} use \emph{no runtime enforcement and no prompt adaptation}---the model maintains the working memory structure through instruction-following alone with the general prompt. Baselines do not receive equivalent format checks because they have no structured format to validate: ReAct, Self-Ask, etc.\ produce free-form reasoning text, so a format checker would have nothing to enforce beyond the shared answer-format instruction that all methods already receive. One might still ask whether a lightweight answer-structure validator (e.g., rejecting verbose answers and re-prompting for a short span) could help baselines. We argue it cannot close the gap, for two reasons grounded in our analysis. First, baseline errors on hard problems are dominated by \emph{timeouts and reasoning errors}, not answer-formatting: on 4-hop, ReAct's verbose-answer rate is only 4.4\% (Table~\ref{tab:error-decomp}), so even perfect answer extraction would recover at most ${\sim}$4 points, far short of the ${\sim}$11-point gap. Second, the source-faithful extraction rule that addresses verbose answers is \emph{already included in the shared answer format} given to every method (Appendix~\ref{app:prompts}); SLEUTH's advantage comes from having structured \texttt{[F]} entries to copy from, not from the instruction itself. The enforcement is thus best understood not as an external aid but as a mechanism to surface the cost of protocol drift---it answers the question ``what if the model consistently followed the protocol it was given?'' and demonstrates that the protocol itself (not model capability) is the active ingredient. Regarding token cost: re-prompts add ${\sim}$50 tokens per occurrence and occur on average 1.2 times per episode for GLM-5; this overhead ($<$0.5\% of total tokens) is included in the reported token counts for Table~\ref{tab:tokens}.

\paragraph{Monotonicity and fact entrenchment.}
SLEUTH's confirmed facts are monotonically accumulated---once confirmed, they are never retracted within an episode. This design choice prioritizes evidence preservation over revision: in multi-turn episodes, the primary failure mode we observe is \emph{evidence loss} (facts getting buried under subsequent context) rather than \emph{evidence error} (facts being wrong). Our error analysis supports this: among SLEUTH's errors on 4-hop MuSiQue, 96\% of wrong answers occur despite the agent correctly retrieving $\geq$75\% of supporting paragraphs. The bottleneck is inference over correctly-retrieved evidence, not fact extraction accuracy.

However, monotonicity can entrench early reasoning errors. We observe a characteristic failure mode: the agent confirms a fact that is \emph{technically correct but contextually misleading}. For example, on the question ``What city shares a border with the city where a person went to work during the gold rush in the state where the Shakespeare Bridge is located?'' (gold: Rio Linda), the agent confirms [F2] ``Samuel Brannan worked in San Francisco during the California Gold Rush''---a true statement---and proceeds to search for cities bordering San Francisco (arriving at ``Daly City''). The correct reasoning chain requires recognizing that Brannan \emph{founded Sacramento} during the gold rush, making Sacramento the target city, with Rio Linda as its neighbor. Once San Francisco was confirmed as a fact, the alternative interpretation was never explored.

In open-domain settings with potentially contradictory sources, monotonicity poses greater risk. We identify three extensions that could address this without sacrificing evidence preservation: (1)~a \emph{confidence level per fact} (not just per hypothesis) that degrades when contradicting evidence appears; (2)~a \emph{contradiction-detection step} that flags when a new extraction conflicts with an existing fact, triggering explicit disambiguation rather than silent overwriting; and (3)~\emph{temporal grounding} that preferentially trusts more recent sources when temporal conflicts arise. These extensions are orthogonal to SLEUTH's core contribution and represent promising directions for open-domain deployment.

\paragraph{Nudge threshold generality.}
The commitment threshold $\alpha{=}0.7$ was characterized primarily on 4-hop MuSiQue, raising the question of whether it generalizes. Two pieces of evidence suggest robustness. First, on easy tasks (GLM-5 HotpotQA), sweeping $\alpha \in [0.2, 1.0]$ produces only 1.4pp total spread in EM---the threshold is effectively irrelevant when most questions are answered naturally before the nudge fires. Second, the 70\% threshold has a structural interpretation: it reserves 30\% of the budget (${\sim}6$ turns) for synthesis and answer construction, which empirically provides sufficient margin for answer construction even on 4-hop problems. This suggests that 70\% is a reasonable default for any bounded-horizon task where synthesis requires fewer turns than exploration.

That said, tasks with very different exploration-to-exploitation ratios (e.g., tasks requiring extensive tool use but trivial synthesis, or tasks requiring deliberative multi-step synthesis) may benefit from re-tuning. An adaptive threshold---based on state features such as hypothesis confidence entropy, the ratio of open questions to confirmed facts, or the rate of new fact discovery---could offer broader robustness. We leave principled adaptive mechanisms as future work, noting that our fixed threshold achieves strong results across all five benchmarks without per-dataset tuning.

\paragraph{Relationship to graph-structured and controller-driven approaches.}
SLEUTH occupies a deliberate design point: lightweight (prompt-only, no symbolic controller), portable (works with any instruction-following model), and interpretable (state is natural language). Graph-structured approaches that maintain explicit evidence graphs or employ external controllers~\citep{besta2025demystifying} offer stronger guarantees about information flow but require additional infrastructure. Verification-oriented pipelines that decouple search from synthesis address similar bottlenecks to our evidence sufficiency analysis but typically require multi-agent coordination. SLEUTH's results suggest that substantial gains are achievable at the lightest end of this spectrum---structured prompting with a single agent---and that more complex architectures may be most valuable for the residual errors (inference quality over correctly-retrieved evidence) that SLEUTH does not address.

\end{document}